\documentclass[runningheads]{llncs}
\usepackage[T1]{fontenc}

\usepackage{graphicx}
\usepackage{amsfonts}
\usepackage{amsmath}

\usepackage{booktabs}
\usepackage{makecell}
\usepackage{subcaption}

\usepackage{multirow}
\usepackage{relsize}

\usepackage{hyperref}
\usepackage{cleveref} 

\usepackage{url}
\usepackage{color}

\hypersetup{
    colorlinks,
    linkcolor={blue},
    citecolor={blue},
    urlcolor={blue}
}

\begin{document}

\title{Template-Based Cortical Surface Reconstruction with Minimal Energy Deformation}

\titlerunning{V2C-MED}

\author{Patrick Madlindl\inst{1} \and
Fabian Bongratz\inst{1,2} \and
Christian Wachinger\inst{1,2}}

\authorrunning{P. Madlindl et al.}

\institute{Lab for AI in Medical Imaging, Technical University of Munich, Munich, Germany
\\
\email{\{patrick.madlindl,fabi.bongratz\}@tum.de}
\and
Munich Center for Machine Learning, Munich, Germany
}

\maketitle
\begin{abstract}
Cortical surface reconstruction (CSR) from magnetic resonance imaging (MRI) is fundamental to neuroimage analysis, enabling morphological studies of the cerebral cortex and functional brain mapping. Recent advances in learning-based CSR have dramatically accelerated processing, allowing for reconstructions through the deformation of anatomical templates within seconds. However, ensuring the learned deformations are optimal in terms of deformation energy and consistent across training runs remains a particular challenge. In this work, we design a Minimal Energy Deformation~(MED) loss, acting as a regularizer on the deformation trajectories and complementing the widely used Chamfer distance in CSR. We incorporate it into the recent V2C-Flow model and demonstrate considerable improvements in previously neglected training consistency and reproducibility without harming reconstruction accuracy and topological correctness.

\end{abstract}
\section{Introduction}
Cortical surface reconstruction~(CSR) is a cornerstone in structural and functional neuroimage analysis~\cite{Esteban2018,Freesurfer}. By extracting the boundaries of cerebral gray matter from magnetic resonance imaging (MRI) and representing them as high-resolution triangular surface meshes, CSR enables topologically faithful modeling of the highly convoluted cortical sheet. It further permits precise morphological measurements, which are critical for investigating normative brain development~\cite{Bethlehem2022} as well as for understanding the onset and progression of neurological and psychiatric disorders~\cite{schizophrenia,AtrophyAlzheimer}. CSR is implemented in established neuroimaging toolboxes such as FreeSurfer~\cite{Freesurfer} and CAT12~\cite{cat12}, for instance. However, a key limitation of these standard approaches is the long runtime in the order of hours, complicating the processing of large databases.

Recently, deep learning-based CSR has emerged as a powerful alternative, offering dramatic reductions in runtime---from hours to seconds. A common approach is to deform a generic input brain template, e.g., a population-average template such as FsAverage~\cite{FreeSurferRegistration}, directly to the individual contours based on deep features extracted from the MRI data~\cite{V2CFlow,CorticalFlow,CorticalFlow++,V2CC,TopoFit}. Compared to segmentation-based~\cite{Henschel2020,CortexODE,le2024diffeomorphic} and implicit CSR~\cite{Gopinath2023,deepcsr}, the template-based approach circumvents costly iso-surface extraction and topology correction since the spherical topology of the cortical ribbon is preserved by construction through the deformation of a topologically correct template mesh (up to potential self-intersections). Moreover, downstream analyses, such as parcellation and group comparisons, are drastically simplified due to the established correspondences to the population templates on which these analyses rely~\cite{V2CFlow}. For training neural networks for CSR, FreeSurfer~\cite{Freesurfer} meshes are commonly employed as a silver-standard reference due to the difficulty of obtaining a reasonable amount of 3D manual delineations of the cortical boundaries. As the reconstruction loss, the Chamfer distance is a popular choice due to its computational efficiency and versatility~\cite{V2CFlow,CorticalFlow,CorticalFlow++,TopoFit}. Importantly, the Chamfer loss does not require point-wise correspondences in the training surfaces. In addition, regularizing loss functions, like the normal consistency or edge loss, ensure regularity and smoothness of the output surfaces~\cite{V2CFlow,CorticalFlow}. 

Yet, despite the recent popularity of deformation-based CSR, two important properties have been largely overlooked in previous studies: the \emph{reproducibility} of training outcomes and the \emph{optimality} of learned deformations. In contrast to traditional CSR, trained CSR models can vary from training to training due to different (usually not reported) initialization and training stochasticity, even when the training set and hyperparameters (usually reported) are kept constant. In contrast to a test-retest evaluation~\cite{Maclaren2014}, where the same model is applied to multiple scans from the same subject acquired (roughly) at the same time, reproducibility concerns the repeatability of training outcomes based on varying model initializations. Ensuring reproducibility is critical for the trustworthiness of CSR models and downstream statistical analyses. 
Optimality, on the other hand, refers to the efficiency of the deformation trajectories learned by the model. Without appropriate constraints, these trajectories may become unnecessarily complex or anatomically implausible, leading to suboptimal reconstructions that may still minimize point-wise losses such as the Chamfer distance. While there exists an optimal-transport approach to CSR based on sliced Wasserstein distances~\cite{le2024diffeomorphic}, it requires a white matter segmentation as the starting point and is thus not applicable to large deformations from a population template. 

In this work, we propose a new regularizing loss function, the \emph{Minimal Energy Deformation~(MED) loss}, for template-based CSR. It is grounded in the intuition that optimal deformations from the template to the reconstruction should encompass minimal deformation energy in terms of the path lengths of warped vertices. Conceptually, the MED loss complements the reconstruction loss, which measures the deviation of the prediction to the reference surface, in that it targets the deformation from the template to the deformed prediction, as illustrated in \Cref{fig:fig-01}. In our experiments, we show that this approach is effective in that it improves the reproducibility and optimality of the predictions of the recently proposed V2C-Flow model~\cite{V2CFlow}.

\section{Methods}
\begin{figure}[t]
    \centering
    \includegraphics[width=0.99\textwidth]{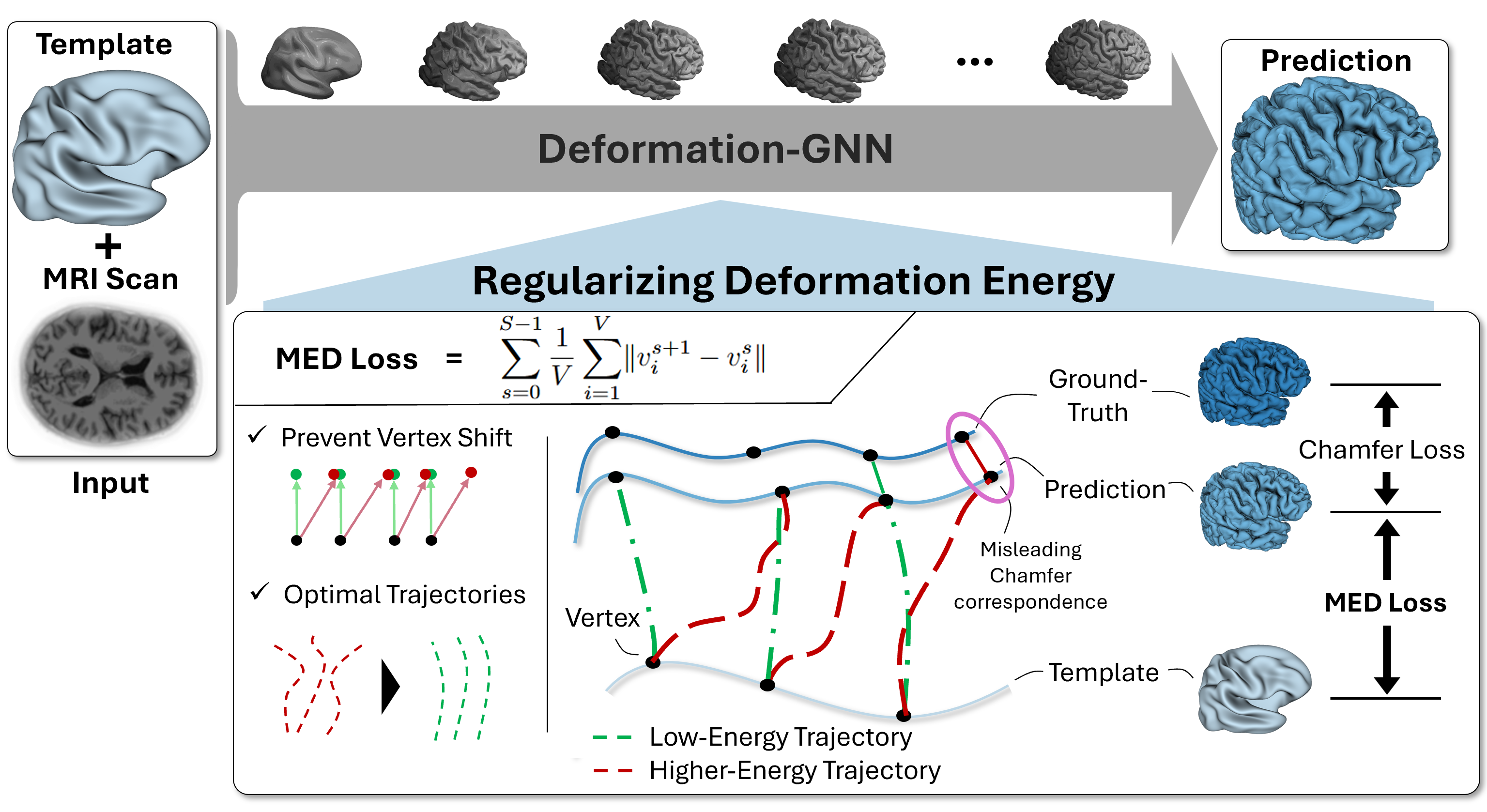}
    \caption{Schematic overview of V2C-MED and the proposed Minimal Energy Deformation (MED) regularization loss. Compared to the standard Chamfer reconstruction loss, which measures the surface discrepancy between prediction and ground truth, the MED loss introduces a novel complementary constraint between prediction and input template. Specifically, the MED loss measures the average path length across $V$ vertices, fostering deformations with minimal deformation energy. The vertex-wise path length is calculated as the sum of $S$ numerical integration steps predicted by the deformation-GNN, which is based on V2C-Flow~\cite{V2CFlow}.
    }
    \label{fig:fig-01}
\end{figure}

We provide a schematic overview of our method, termed Vox2Cortex with Minimal Energy Deformation~(V2C-MED) in \Cref{fig:fig-01}. 

\subsection{V2C-Flow}
V2C-MED builds upon Vox2Cortex-Flow~(V2C-Flow)~\cite{V2CFlow}, which predicts a dynamic vertex-wise deformation field to deform the input mesh template to the 3D brain contours visible in the input MRI scan. The deformation field is integrated with a forward Euler integration scheme, comprising $S=10$ integration steps in total. In a nutshell, V2C-Flow leverages a graph neural network~(GNN) to operate on the surface meshes and a 3D convolutional neural network~(CNN) for deep feature extraction from the MRI scan. It is trained on a sum of curvature-weighted Chamfer loss $\mathcal{L}_{Ch}$, mesh edge loss $\mathcal{L}_e$, normal consistency loss $\mathcal{L}_{nc}$, and cross-entropy segmentation loss $\mathcal{L}_{ce}$. Specifically, the original V2C-Flow loss is given as
\begin{equation} \label{eq:v2c-flow}
    \mathcal{L}_{\text{V2C-Flow}} = \mathcal{L}_{Ch} + \mathcal{L}_{e} + 0.001 \, \mathcal{L}_{nc} + \mathcal{L}_{ce},
\end{equation}
where the relative weighting of the loss terms adheres to the original configuration proposed in V2C-Flow.
Moreover, V2C-Flow connects inner (white matter) and outer (pial) cortical surfaces via virtual edges and deforms them simultaneously for coherent outputs. For architectural and training details, we refer to the original V2C-Flow paper~\cite{V2CFlow}.

\subsection{Minimal Energy Deformation~(MED) Loss}
The motivation for our Minimal Energy Deformation~(MED) loss stems from the fact that V2C-Flow, similar to other related CSR methods~\cite{V2CC,CorticalFlow,CorticalFlow++}, is trained solely on the produced predictions, using a Chamfer reconstruction loss, without considering \emph{how} these predictions were obtained, i.e., the actual deformation trajectories. Specifically, a vertex shift due to suboptimal network initialization and misleading closest-point distances between predicted and ground-truth surfaces can easily occur in the Chamfer loss. Moreover, the smoothness regularization with edge and normal consistency losses does not help in this regard.
At the same time, optimal deformations in terms of minimal deformation energy are essential for stable, reproducible, and reliable model predictions. 

As a solution, we propose to regularize the training loss based on the mean L2 path lengths of all ($V$) template vertices. The path lengths are obtained from the sum over $S$ numerical integration steps and accumulated across white matter~(W) and pial~(P) boundaries, yielding the following training loss function:
\begin{equation}
\mathcal{L}_{MED} = \sum_{\substack{c\in  \{\text{W,P}}\}} \sum_{s=0}^{S-1}  \frac{1}{V} \sum_{i=1}^V \lVert v_{c,i}^{s+1} - v_{c,i}^s \rVert_2.
\label{eq: E-loss}
\end{equation}
In contrast to optimal transport/Wasserstein distance-based loss functions~\cite{le2024diffeomorphic}, which require a costly search of the optimal transport plan between prediction and ground-truth shapes, we exploit the point-wise vertex correspondence between the template and the deformed shapes in \Cref{eq: E-loss}. By treating every vertex equally, i.e., giving each vertex a unit weight, $\mathcal{L}_{MED}$ provides a measure of global deformation energy. During training, we add it to the original V2C-Flow loss, yielding
\begin{equation}
    \mathcal{L}_{\text{V2C-MED}} = \mathcal{L}_{\text{V2C-Flow}} + \lambda \, \mathcal{L}_{MED}
\end{equation}
as the loss function for our V2C-MED model. Tuning of the regularization weight $\lambda$, based on our CSR validation set, suggests that setting $\lambda=0.01$ is a reasonable choice in practice.

\subsection{Evaluation of Reproducibility}\label{sec:reprod}
Previously, CSR models were evaluated regarding surface accuracy, topological correctness, and test-retest reliability using established metrics~\cite{V2CFlow,Gopinath2023,CorticalFlow,CortexODE}. However, the reproducibility of training outcomes with different model initializations has been widely ignored. To this end, we trained three models with random seeds $\tau_0, \tau_1, \tau_2$, affecting the initialization of both the GNN and the CNN model weights. In V2C-Flow and V2C-MED, we initialized the GNN weights using a standard normal distribution, except for the output layers, which we always initialized with zeros for the sake of training stability~\cite{V2CFlow}. For the CNN, we use the standard Kaiming uniform initialization~\cite{kaiming15} as implemented in PyTorch~(v1.10). 

As a metric for reproducibility, we compute the root mean square deviation~(RMSD) per vertex $v$ from its mean $\bar{v}$ across the three runs. Let $v(\tau_i)$ be a predicted vertex in the model initialized with seed $\tau_i$, then our vertex-wise RMSD is given as
\begin{equation}
\text{RMSD}(v) = \sqrt{\frac{1}{3}\sum_{i=0}^{2} \lVert v(\tau_i) - \bar{v}\rVert_2^2}.
\label{eq: RMSD}
\end{equation}
Finally, a single reproducibility score is obtained by aggregating the mean across all vertices and subjects in the test set.

\subsection{Implementation and Experimental Setup}
Our implementation of V2C-MED builds upon the official V2C-Flow repository, which is publicly available at \url{https://github.com/ai-med/Vox2Cortex}. To ensure a fair and controlled comparison, we adopted the same training protocol as described in~\cite{V2CFlow}, including data preprocessing, network architecture, and optimization settings. This controlled setup allows us to isolate and evaluate the specific contribution of the proposed Minimal Energy Deformation (MED) regularization loss. All experiments were conducted under identical conditions to highlight the impact of the MED loss on training stability, reproducibility, and reconstruction quality. Due to its popularity for downstream applications, we used the FsAverage template~\cite{FreeSurferRegistration}, respectively a smoothed version (called V2C-Flow-S in~\cite{V2CFlow}) as the input to all models. We considered two different training template resolutions: ico-6 (40,962 vertices per surface) and ico-7 (163,842 vertices per surface) tessellation; at test time, we always used the highest possible template resolution~(ico-7). The surface up-scaling at test time has previously been shown to be beneficial in cases where GPUs with more than 24 GB of VRAM are not available~\cite{V2C}. Additionally, for the sake of feasibility (we needed to train three models per configuration from scratch with different random seeds to assess reproducibility, cf.~\Cref{sec:reprod}), we focus on the right brain hemisphere here. By cutting the original scans in half in the sagittal plane, we obtained input scans of dimension $[96\times208\times192]$. We used a maximum number of 85 training epochs per model, which we found to be sufficient for convergence in this setting. This corresponds to around two days of training on a single Nvidia A100 GPU with the ico-6 tessellation; for the ico-7 templates, the training took around four days per model. We consistently picked the model with the lowest reconstruction error based on the validation score for the final test-set evaluation.

\subsection{Datasets}
We used data from the Alzheimer's Neuroimaging Initiative (ADNI) (\url{http://adni.loni.usc.edu/}), which comprises T1-w MRI scans~(1mm isotropic resolution, registered to MNI152 standard space) of subjects diagnosed with Alzheimer's Disease, mild cognitive impairment, and cognitively normal condition. We only used baseline scans, split into 1154 training, 169 validation, and 323 test cases. We balanced the splits with respect to diagnosis, sex, and age. Moreover, we use a test-retest~(TRT) database~\cite{Maclaren2014}, comprising 40 scans from 3 subjects, respectively, to evaluate the reliability of reconstructions across scans. As a silver-standard reference for the reconstruction accuracy, we used FreeSurfer (v7.2)~\cite{Freesurfer}. Note, however, that the evaluation of the training reproducibility is independent of FreeSurfer.

\section{Results and Discussion}

\begin{table}[t]
\caption{We report reconstruction accuracy based on our test set in terms of average symmetric surface distance (ASSD), topological correctness in terms of percentage of self-intersecting faces (\%SIF), test-retest reliability (TRT), the deformation energy in normalized image space, and the reproducibility in terms of RMSD as defined in \Cref{sec:reprod}. All values are mean\textpm SD across the three training runs except for RMSD (mean\textpm SD across the mesh vertices). 
Lower is better for all metrics; best values are highlighted.
}\label{tab:Results-All}
\centering
\setlength{\tabcolsep}{3pt}
\renewcommand\bfdefault{b}
\begin{tabular}{l l c c c c}
    \toprule
    & & \multicolumn{2}{c}{Train ico-6} & \multicolumn{2}{c}{Train ico-7} \\
     \cmidrule(lr){3-4}
     \cmidrule(lr){5-6}
    Metric & Surf. & V2C-Flow & V2C-MED & V2C-Flow & V2C-MED \\
    \midrule
    \multirow{2}{*}{Accuracy (ASSD, mm)} 
        & WM & 0.285\textpm0.005 & 0.292\textpm0.010 & \textbf{0.204\textpm0.007} & 0.217\textpm0.001 \\
        & Pial & 0.293\textpm0.003 & 0.297\textpm0.002 & \textbf{0.202\textpm0.003} & 0.207\textpm0.002\\
    \midrule
    \multirow{2}{*}{\%SIF}
        & WM    & 1.326\textpm0.116 & \textbf{0.741\textpm0.033} & 1.766\textpm0.733 & 2.250\textpm0.303 \\
        & Pial  & 2.519\textpm0.283 & \textbf{1.476\textpm0.211} & 2.629\textpm0.597 & 2.531\textpm0.437 \\
    \midrule
     \multirow{2}{*}{Reliability (TRT, mm)}
        & WM    & 1.298\textpm0.030 & 1.211\textpm0.028 & 1.229\textpm0.030 & \textbf{1.116\textpm0.029} \\
         & Pial & 1.404\textpm0.039 & 1.323\textpm0.032 & 1.312\textpm0.030 & \textbf{1.202\textpm0.024} \\
    \midrule
    \multirow{2}{*}{\makecell[l]{Deformation Energy}}
        & WM   & 0.164\textpm0.008 & 0.084\textpm0.001 & 0.151\textpm0.017 & \textbf{0.081\textpm0.001} \\
        & Pial & 0.176\textpm0.009 & 0.094\textpm0.001 & 0.162\textpm0.017 & \textbf{0.091\textpm0.001} \\
          \midrule
   
    \multirow{2}{*}{\makecell[l]{Reproducibility\\ (RMSD, mm)}}
        & WM  & 1.827\textpm0.415 & 1.348\textpm0.388 & 2.548\textpm2.182 &\textbf{1.290\textpm0.962} \\
        & Pial & 1.919\textpm0.423 & 1.436\textpm0.413 & 2.598\textpm2.121 & \textbf{1.326\textpm0.880} \\
   
    \bottomrule
\end{tabular}
\end{table}

We report the performance of all trained models on the test set, as summarized in \Cref{tab:Results-All}. The reported standard deviation (SD) reflects variability across the three independent training runs for each model. An exception is made for RMSD, which---by design---aggregates results across all three runs; in this case, the SD is computed across the mesh vertices.

\begin{figure}[t]
    \centering
    \includegraphics[width=0.95\textwidth]{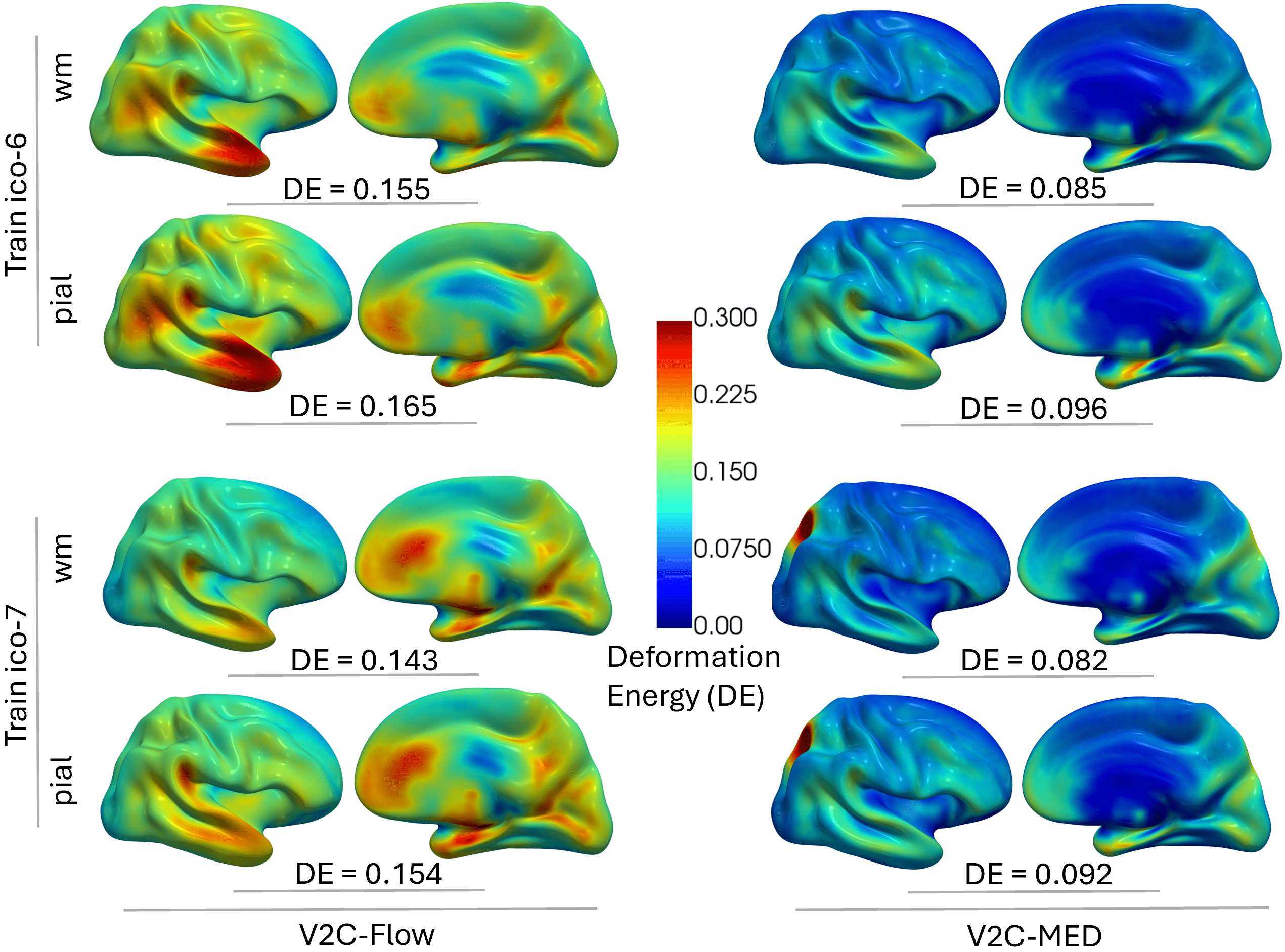}
    \caption{The vertex-wise mean deformation energy of white matter (wm) and pial surfaces on our test set in ico-6 (top) and ico-7 (bottom) training runs for V2C-Flow (left) and V2C-MED (right). Lower values indicate lower-energy deformations. The visualization is based on the smoothed FsAverage template.}
    \label{fig:deform energies}
\end{figure}

First, as a sanity check, we observe that the proposed MED loss is effective in that it reduces the deformation energy by around 50-60\% in all scenarios (see also \Cref{fig:deform energies}). At the same time, the reconstruction accuracy remains particularly stable. Although the best accuracy is obtained with the original V2C-Flow model, trained on the ico-7 tessellation, the advantage over V2C-MED is small, with the average difference in accuracy between V2C-Flow and V2C-MED being less than 0.015 mm. The resolution of the training template, on the other hand, has a much larger impact on the model accuracy ($\sim$0.2 mm for ico-7 vs.~$\sim$0.3 mm for ico-6 training), as expected. 
In terms of the reproducibility of the surface accuracy, all models perform relatively well, as evidenced by the low standard deviation in average symmetric surface distance (ASSD) across runs. Notably, V2C-MED shows a slight advantage in most cases, except for the white matter (WM) surfaces trained on the ico-6 resolution. The RMSD, however, reveals significant differences between V2C-Flow and V2C-MED: the MED loss reduces the RMSD by around 25\% when training with the ico-6 resolution, and up to around 50\% for the ico-7 training. This suggests that while V2C-Flow maintains consistent reconstruction accuracy across runs, the underlying deformation patterns vary significantly---highlighting that \emph{how} the model achieves a certain accuracy is less stable without the MED regularization.
Finally, we found a positive effect of the MED loss on the ratio of self-intersecting faces when training with the ico-6 template, and on the test-retest reliability.
In summary, these results demonstrate consistent quantitative improvements by V2C-MED in reproducibility and deformation energy while maintaining high reconstruction fidelity. Moreover, the consistent improvements across different mesh resolutions suggest that the MED loss generalizes well and can be effectively integrated into various training setups.

\begin{figure}[t]
    \centering
    \includegraphics[width=0.95\textwidth]{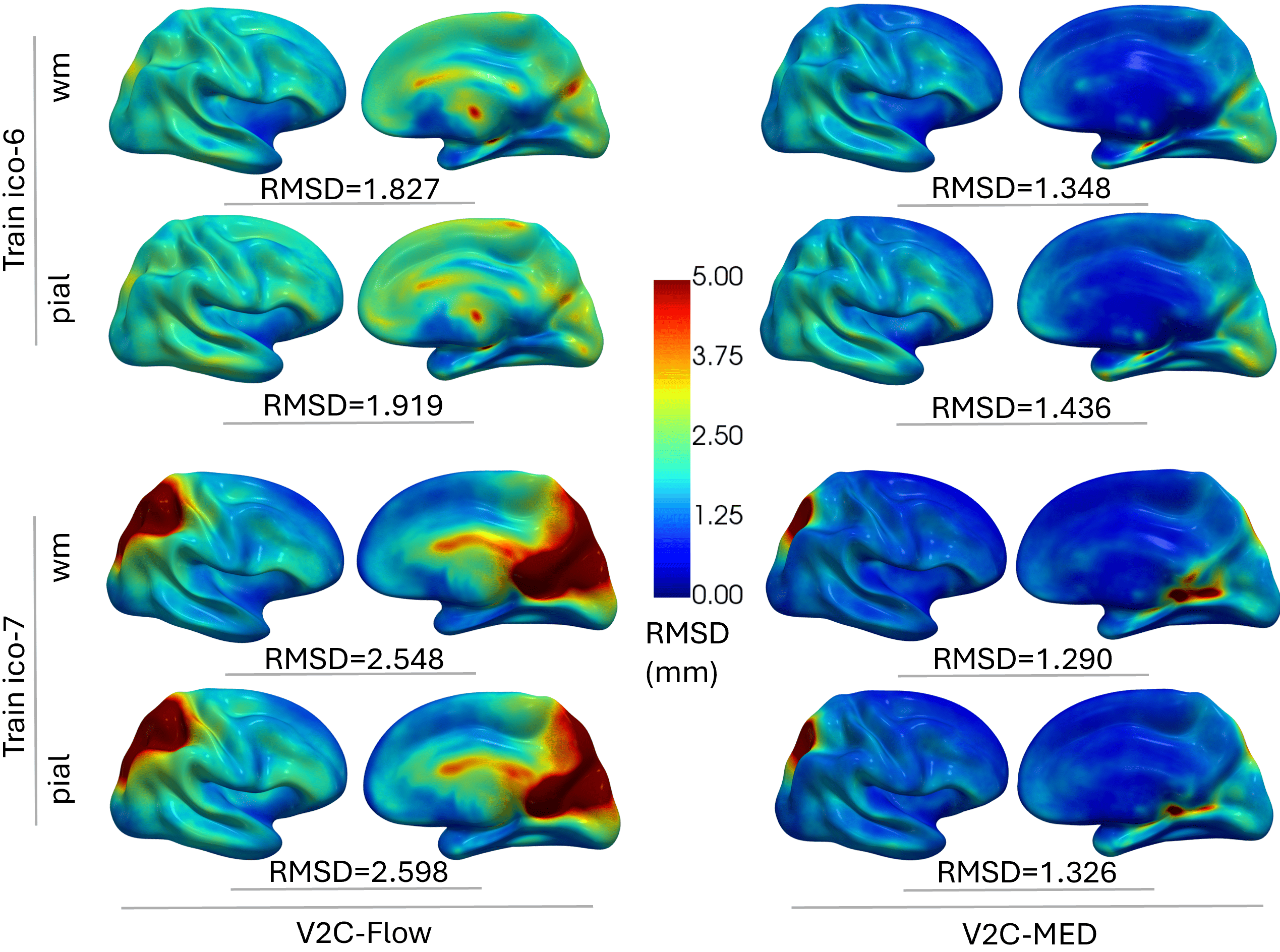}
    \caption{The vertex-wise reproducibility (RMSD) of wm and pial surfaces on our test set in ico-6 (top) and ico-7 (bottom) trainings for V2C-Flow (left) and V2C-MED (right). Lower values indicate better reproducibility. The visualization is based on the smoothed FsAverage template.}
    \label{fig:ADNI RMSD}
\end{figure}

To obtain further insights into the local reproducibility of CSR, we plot the RMSD per vertex in \Cref{fig:ADNI RMSD}. These visualizations complement the findings from \Cref{tab:Results-All}; they show that the improvements in reproducibility achieved by the MED loss are consistent across the entire cortical sheet. Nevertheless, considerable variation in vertex placement remains in the lateral parietal cortex and in the lingual gyrus.

\section{Conclusion}
We introduced the Minimal Energy Deformation (MED) loss and integrated it into V2C-Flow for cortical surface reconstruction. The resulting cortical surface reconstruction method, named V2C-MED, demonstrated significant improvements in terms of deformation energy required to warp the input template to the predicted surfaces. In contrast to previous studies, we conducted multiple trainings with varying random seeds to evaluate the inter-training reproducibility and consistency of predictions across models with different initializations. Our results showed that the MED loss fosters higher reproducibility and consistency across different training scenarios. These findings highlight the potential of the MED loss as a robust regularization strategy for enhancing the stability and trustworthiness of template-based cortical surface reconstruction models.

\begin{credits}
\subsubsection{\ackname} This research was partially supported by the German Research Foundation (DFG, No. 460880779). We gratefully acknowledge the computational resources provided by the Leibniz Supercomputing Centre (www.lrz.de).

\end{credits}

\bibliographystyle{splncs04}
\bibliography{bibliography.bib}

\end{document}